\documentclass[11pt]{article} % For LaTeX2e
\usepackage{jmlr2e}

\usepackage[utf8]{inputenc} % allow utf-8 input
\usepackage[T1]{fontenc}    % use 8-bit T1 fonts
\usepackage{hyperref}       % hyperlinks
\hypersetup{
    colorlinks=true,
    linkcolor=blue,
    citecolor=blue,
    urlcolor=blue,
}
\usepackage{enumitem}
\usepackage{bm}
\usepackage{url}            % simple URL typesetting
\usepackage{wrapfig,lipsum,booktabs}       % professional-quality tables
\usepackage{amsfonts}       % blackboard math symbols
\usepackage{nicefrac}       % compact symbols for 1/2, etc.
\usepackage{microtype}      % microtypography

\usepackage{caption}
\usepackage{subcaption}

\usepackage{amssymb}
\usepackage{algorithm}
\usepackage{algorithmicx}
\usepackage{framed,graphicx}
\usepackage{xcolor}
\usepackage{amsmath}
\usepackage{bigints}
\usepackage{multirow}
\usepackage{pifont}
\usepackage{xspace}
\newcommand{\xmark}{\ding{55}\xspace}%

\firstpageno{1}

\begin{document}

\title{\textsc{Mega}: Moving Average Equipped Gated Attention}

\author{\name Xuezhe Ma\thanks{Equal Contribution. Correspondence to xuezhema@isi.edu and chuntinz@fb.com} \email xuezhema@isi.edu \\
\addr University of Southern California  \\
\name Chunting Zhou$*$ \email chuntinz@fb.com \\
\addr Meta AI Research \\
\name Xiang Kong \email xiangk@cs.cmu.edu \\
\addr Carnegie Mellon University \\
\name Junxian He \email junxianh2@gmail.com \\
\addr Shanghai Jiao Tong University \\
\name Liangke Gui \email liangkeg@cs.cmu.edu \\
\name Graham Neubig \email gneubig@cs.cmu.edu \\
\addr Carnegie Mellon University \\
\name Jonathan May \email jonmay@isi.edu \\
\addr University of Southern California  \\
\name Luke Zettlemoyer \email lsz@fb.com \\
\addr Meta AI Research \\
}
\editor{}

\maketitle

\begin{abstract}
The design choices in the Transformer attention mechanism, including weak inductive bias and quadratic computational complexity, have limited its application for modeling long sequences. 
In this paper, we introduce \textsc{Mega}, a simple, theoretically grounded, single-head gated attention mechanism equipped with (exponential) moving average to incorporate inductive bias of position-aware local dependencies into the position-agnostic attention mechanism. 
We further propose a variant of \textsc{Mega} that offers linear time and space complexity yet yields only minimal quality loss, by efficiently splitting the whole sequence into multiple chunks with fixed length.
Extensive experiments on a wide range of sequence modeling benchmarks, including the Long Range Arena, neural machine translation, auto-regressive language modeling, and image and speech classification, show that \textsc{Mega} achieves significant improvements over other sequence models, including variants of Transformers and recent state space models.\footnote{The implementation of the algorithm is available at \url{https://github.com/facebookresearch/mega}}
\end{abstract}

\section{Introduction}
Designing a single unified model to capture long range dependencies in sequential data across a diverge range of modalities, such as language, audio, image and video, is a central and challenging problem in sequence modeling.
A number of different archtectures have been developed, including convolutional neural networks (CNNs)~\citep{kim-2014-convolutional,strubell-etal-2017-fast}, recurrent neural networks (RNNs)~\citep{goller1996learning,hochreiter1997long,cho2014properties}, Transformers~\citep{vaswani2017attention} and recent state space models (SSMs)~\citep{gu2022efficiently,mehta2022long}.
Among these models, the Transformer architecture~\citep{vaswani2017attention} has stood out for its impressive empirical success on a wide range of language and vision tasks, including machine translation~\citep{vaswani2017attention,ott2018scaling}, language understanding~\citep{devlin2019bert,liu2019roberta}, image recognition~\citep{dosovitskiy2020image,touvron2021training} and genetic sequence modeling~\citep{madani2020progen,jumper2021highly}, mainly because of the conceptually attractive attention mechanism~\citep{bahdanau2015neural,luong2015effective,vaswani2017attention} which directly models interactions between each pair of input tokens. 

\begin{table}
\centering
\caption{Experimental results of Transformer (XFM),  S4 and \textsc{Mega} on five sequence modeling benchmarks of different types of data, including long range arena (LRA), machine translation (WMT16 en-de), language modeling (WikiText-103), image classification (ImageNet-1k), raw speech classification (SC-Raw). }
\label{tab:main_results}
\resizebox{0.99\columnwidth}{!}{
\begin{tabular}{l|ccccc}
\toprule
 & \textbf{LRA} (Acc. $\uparrow$) & \textbf{WMT16} (BLEU $\uparrow$) & \textbf{WT103} (PPL. $\downarrow$) & \textbf{ImageNet} (Acc. $\uparrow$) & \textbf{SC} (Acc. $\uparrow$) \\
\midrule
XFM & 59.24 & 27.97 & 18.66 & 81.80 & \xmark \\
S4 & 85.86 &  --  & 20.95 &  -- & \textbf{97.50} \\
\textsc{Mega} & \textbf{88.21} & \textbf{29.18} & \textbf{18.07} & \textbf{82.31} & 97.30 \\
\bottomrule
\end{tabular}
}
\end{table}

Attention provides the key mechanism that captures contextual information from the entire sequence by modeling pairwise interactions between the inputs at every timestep. 
However, there are two common drawbacks in the design of attention mechanism: i) \emph{weak inductive bias}; and ii) \emph{quadratic computational complexity}.
First, the attention mechanism does not assume prior knowledge of the patterns of dependencies between tokens (e.g. positional inductive bias), instead learning to predict the pairwise attention weights directly from data.
Second, the cost to compute and store the attention weights is quadratic in the length of the input sequences.
Recent studies have shown the limitations of applying Transformers to long sequence tasks,  w.r.t both accuracy and efficiency~\citep{tay2020efficient}.

\begin{figure}[t]
\centering
\includegraphics[width=0.99\textwidth]{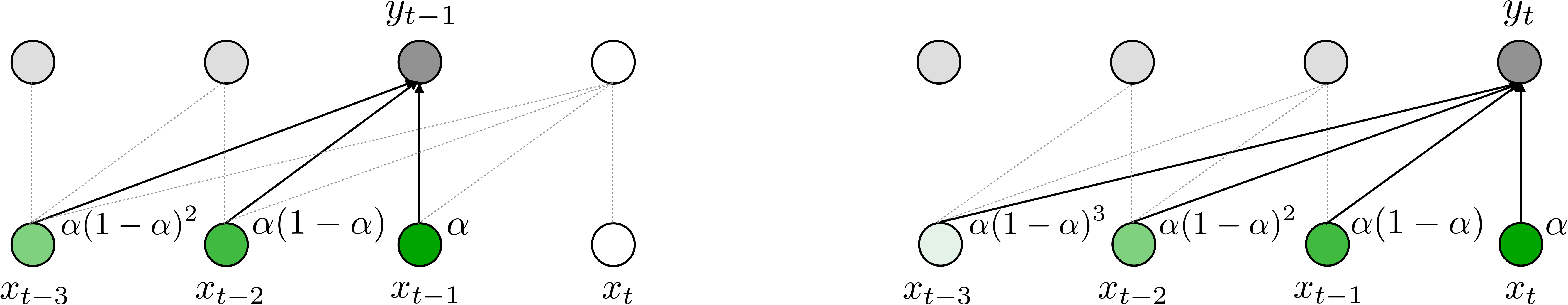}
\caption{Illustration of the exponential moving average (EMA) approach, which averages the input values $\boldsymbol{X}$ with weights decaying exponentially over timesteps.}
\label{fig:ema}
\vspace{-3mm}
\end{figure}

In this work, we propose a \emph{moving average equipped gated attention mechanism} (\textsc{Mega}) to solve the two weaknesses simultaneously. 
The key idea is to incorporate inductive biases into the attention mechanism across the timestep dimension, by leveraging the classic  exponential moving average (EMA) approach~\citep{hunter1986exponentially}.
EMA captures local dependencies that exponentially decay over time (see Figure~\ref{fig:ema}), and has been widely used in time series data modeling (\S\ref{sec:background}).
We introduce a multi-dimensional damped form of EMA with learnable coefficients (\S\ref{subsec:mddema}), and subsequently develop the moving average equipped gated attention mechanism by integrating the EMA with a variant of the single-head gated attention~\citep{hua2022transformer} (\S\ref{subsec:mega}).
Theoretically, we show that the single-head gated attention is as expressive as the most commonly used multi-head attention (\S\ref{subsec:theory}). 
Benefiting from the incorporated moving average mechanism, we further propose a variant of \textsc{Mega} with linear complexity, named \textsc{Mega}-chunk,  which simply chunks input sequences into fixed blocks with minimal loss of contextual information (\S\ref{subsec:mega-chunk}).

Experimentally, through five sequence modeling tasks across various data types, including long-context sequence modeling, neural machine translation, auto-regressive language modeling, and image and speech classification, we demonstrate that \textsc{Mega} significantly outperforms a variety of strong baseline models, in terms of both effectiveness and efficiency~(\S\ref{sec:experiments}) (see Table~\ref{tab:main_results}). 
These improvements illustrate the importance of modeling long- and short-term dependencies via different patterns of inductive biases.

\section{Background}
\label{sec:background}
In this section, we set up notations, briefly review two widely used approaches for sequence modeling---the self-attention mechanism and exponential moving average (EMA)---and discuss the motivation for combining them.

We use $\boldsymbol{X} = \{\mathbf{x}_1, \mathbf{x}_2, \ldots, \mathbf{x}_n\} \in \mathbb{R}^{n\times d}$ to denote a sequence of input representations with length $n$.
%where $\mathbf{x}_t \in \mathbb{R}^{d}$ is the $d$-dimensional representation vector of the $t$-th token of $\boldsymbol{X}$ with $t \in \{1, 2, \ldots, n\}$.
Let $\boldsymbol{Y} =\{\mathbf{y}_1, \mathbf{y}_2, \ldots, \mathbf{y}_n\} \in \mathbb{R}^{n\times d}$ be the sequence of output representations of each layer with the same length $n$ as the input $\boldsymbol{X}$.
%The goal of a sequence model is to compute $\boldsymbol{Y}$ to capture the contextual information from $\boldsymbol{X}$ necessary to solve the target task.
In this paper, we assume the representations of the input and output sequences have the same dimension $d$. 

\subsection{Self-Attention Mechanism}
\label{subsec:attention}
%\gn{On an auxiliary note, do you want to explain GAU in equations here? GAU is important later in the paper, and it might be good to give the readers background into it.}
%Replay by Chuntingz: in section 2, we want to highlight the properties and limitations of EMA and attention to motivate the combination of them. Thus, we do not want to introduce GAU here.

The traditional self-attention mechanism is a function:
\begin{equation}\label{eq:attention}
\boldsymbol{Y} = \mathrm{Attn}(\boldsymbol{X}) = f\left( \frac{\boldsymbol{Q} \boldsymbol{K}^T}{\tau(\boldsymbol{X})} \right) \boldsymbol{V}
\end{equation}
where $\mathrm{Attn}: \mathbb{R}^{n\times d} \rightarrow \mathbb{R}^{n\times d}$ is the self-attention function.
$\boldsymbol{Q}=\boldsymbol{X}W_q + b_q, \,\, \boldsymbol{K} = \boldsymbol{X}W_k + b_k, \,\, \mbox{and~} \boldsymbol{V} = \boldsymbol{X}W_v + b_v$ are the sequences of queries, keys and values, with learnable parameters $W_q, \,\, W_k, \,\, W_v \in \mathbb{R}^{d\times d}$, and $b_q, \,\, b_k, \,\, b_v \in \mathbb{R}^{d}$. $f(\cdot)$ is an attention function, e.g. the softmax function $f_{\mathrm{softmax}}(\cdot)$~\citep{bahdanau2015neural}, or the recently proposed squared ReLU function $f_{\mathrm{relu^2}}(\cdot)$~\citep{so2021searching,hua2022transformer}.
$\tau(\boldsymbol{X})$ is a scaling term, which is commonly set to $\tau(\boldsymbol{X}) = \sqrt{d}$ for $f_{\mathrm{softmax}}(\cdot)$, or $\tau(\boldsymbol{X}) = n$ for $f_{\mathrm{relu^2}}(\cdot)$.
The commonly used multi-head variant of attention performs the attention function $h$ times in parallel.

We can define a matrix $\boldsymbol{A}=f(\frac{\boldsymbol{Q}{\boldsymbol{K}}^{T}}{\tau(\boldsymbol{X})})\in \mathbb{R}^{n\times n}$ following~\eqref{eq:attention}, which is called the \textit{attention matrix}, as it specifies the weight of the dependency strength between every pair of tokens in $\boldsymbol{X}$.
Since it models pairwise dependency weights, the matrix $\boldsymbol{A}$ in principle delivers a flexible and powerful mechanism to learn long-distance dependencies with minimal inductive biases. 
However, it is in practice a challenging task to recognize all the dependency patterns in $\boldsymbol{A}$ directly from data, particularly when processing long sequences.
Moreover, calculating $\boldsymbol{A}$ with $h$ attention heads takes $O(hn^2)$ time and space, and the  quadratic dependency on sequence length becomes a significant bottleneck.

\subsection{Exponential Moving Average (EMA)}
\label{subsec:ema}
The moving average is a classic approach for sequential data modeling, which has been widely used in time series data to smooth out short-term fluctuations and highlight long-term trends or cycles.
The Exponential Moving Average (EMA)~\citep{winters1960forecasting,hunter1986exponentially}, a special case of moving average, applies weighting factors that decrease exponentially. 
Formally, an EMA recursively calculates the output sequence $\boldsymbol{Y}$:
\begin{equation}
\label{eq:ema}
\mathbf{y}_t = \boldsymbol{\alpha} \odot \mathbf{x}_t + (1 - \boldsymbol{\alpha}) \odot \mathbf{y}_{t-1},
\end{equation}
where $\boldsymbol{\alpha} \in (0, 1)^{d}$ is the EMA coefficient representing the degree of weighting decrease, and $\odot$ is the element-wise product.
A higher $\boldsymbol{\alpha}$ discounts older observations faster (see Figure~\ref{fig:ema}).
%The graphical specification of EMA is illustrated in Figure~\ref{fig:ema}.

Using an EMA places a strong  inductive bias on the learning of pairwise dependencies: the dependency weight between two tokens decreases exponentially over time with an input-agnostic decay factor $\boldsymbol{\alpha}$.
This property favors local dependencies, and limits long-distance dependencies.
Despite the recurrent formulation in \eqref{eq:ema}, the computation of EMA can be represented as $n$ individual convolutions, which can be computed efficiently using fast Fourier transforms (FFTs) (see Appendix~\ref{appendix:ema} for details).

\subsection{Why Combine Attention with EMA?}
\label{subsec:problem}

As discussed in Sections~\ref{subsec:attention} and \ref{subsec:ema}, EMA and attention mechanisms each have their own limitations, despite their wide applications and impressive successes in sequence modeling.
%In this paper, we propose a simple and effective neural architecture to simultaneously tackle the aforementioned challenges of the EMA and attention mechanism, by leveraging their properties to complement each other.
By leveraging their properties to complement each other, we propose to embed an EMA into the calculation of the attention matrix $\boldsymbol{A}$. The resulting model enjoys the benefit from strong inductive bias, while maintaining the capacity to learn complex dependency patterns.
Moreover, this integration enables the design of a computationally efficient chunk-wise attention mechanism with linear complexity w.r.t sequence length (\S\ref{subsec:mega-chunk}).

\section{Moving Average Equipped Gated Attention (\textsc{Mega})}
\label{sec:mega}
In this section, we describe in detail our proposed method, \emph{moving average equipped gated attention} (\textsc{Mega}).
We first introduce multi-dimensional damped EMA (\S\ref{subsec:mddema}), which is a key component combined with the single-head gated attention in \textsc{Mega} (\S\ref{subsec:mega}), and discuss the relationship between \textsc{Mega} and three closely related models: GRU~\citep{cho2014properties}, Flash~\citep{hua2022transformer} and S4~\citep{gu2022efficiently}.
We also provide theoretical justification for the design of single-head gated attention  (\S\ref{subsec:theory}).
Then, we describe the detailed architecture of each \textsc{Mega} block, including feed-forward and normalization layers (\S\ref{subsec:mega_layer}). 
At last, we present \textsc{Mega}-chunk, a variant of \textsc{Mega} that simply splits input sequences into fixed chunks, reducing time and space complexity from quadratic to linear (\S\ref{subsec:mega-chunk}).

\subsection{Multi-dimensional Damped EMA}
\label{subsec:mddema}
\textsc{Mega} introduces a modification of the standard EMA, named \emph{multi-dimensional damped EMA}, to improve its flexibility and capacity.

\paragraph{Damped EMA.} 
Previous studies~\citep{mckenzie2010damped,svetunkov2016complex} have shown that relaxing the coupled weights of the previous and current observations  ($\boldsymbol{\alpha}$ vs. $1 - \boldsymbol{\alpha}$ in \eqref{eq:ema}) produces robust dependency modeling.
Inspired by this, \textsc{Mega} allows the damping of the influence of the previous time step:
\begin{equation}
\label{eq:damping}
\mathbf{y}_t = \boldsymbol{\alpha} \odot \mathbf{x}_t + (1 - \boldsymbol{\alpha} \odot \boldsymbol{\delta}) \odot \mathbf{y}_{t-1},
\end{equation}
where $\boldsymbol{\delta} \in (0, 1)^{d}$ is the damping factor. 

\paragraph{Multi-dimensional Damped EMA.}
To further improve the expressiveness of EMA, we introduce a multi-dimensional variant of  EMA. 
Concretely, we first expand each dimension of the input sequence $\boldsymbol{X}$ individually into $h$ dimensions via an expansion matrix $\boldsymbol{\beta} \in \mathbb{R}^{d\times h}$. 
Formally, for each dimension $j \in \{1, 2, \ldots, d\}$:
\begin{equation}
\mathbf{u}^{(j)}_t = \boldsymbol{\beta}_j \mathbf{x}_{t,j}
\end{equation}
where $\boldsymbol{\beta}_j \in \mathbb{R}^{h}$ is the $j$-th row of $\boldsymbol{\beta}$,  $\mathbf{u}^{(j)}_t \in \mathbb{R}^{h}$ is the expanded $h$-dimensional vector for the $j$-th dimension at timestep $t$. 
%\gn{I found this explanation confusing, aren't these just the equations for a vector outer product? If you're doing something different than an outer product it'd be good if you could explain what the difference is. Also, is $\mathbf{u}^{(j)}_t \in \mathbb{R}^{h}$ correct? It seems that it should be $\mathbf{u}_t \in \mathbb{R}^{h}$}
% reply by Max: It is not vector outer product. and $\mathbf{u}^{(j)}_t \in \mathbb{R}^{h}$ is correct. For each $\mathbf{x}_{t,j}$, we expand it to $h$ dimensions by a different vector $\beta$_j.

Correspondingly, we extend the shape of $\boldsymbol{\alpha}$ and $\boldsymbol{\delta}$ from a one-dimensional vector to a two-dimensional matrix, i.e. $\boldsymbol{\alpha}$, $\boldsymbol{\delta} \in \mathbb{R}^{d\times h}$, where $\boldsymbol{\alpha}_j$, $\boldsymbol{\delta}_j \in \mathbb{R}^{h}$ denote the $j$-th row of $\boldsymbol{\alpha}$ and $\boldsymbol{\delta}$, respectively. 
Then, for each dimension $j$, the damped EMA is applied to the $h$-dimensional hidden space: 
\begin{align}
\label{eq:mddema}
\mathbf{h}^{(j)}_t & = \boldsymbol{\alpha}_j \odot \mathbf{u}^{(j)}_t + (1 - \boldsymbol{\alpha}_j \odot \boldsymbol{\delta}_j) \odot \mathbf{h}^{(j)}_{t-1} \nonumber \\
\mathbf{y}_{t,j} & = \boldsymbol{\eta}^T_j \mathbf{h}^{(j)}_t
\end{align}
where $\mathbf{h}^{(j)}_t \in \mathbb{R}^{h}$ is the EMA hidden state for the $j$-th dimension at timestep $t$. 
$\boldsymbol{\eta} \in \mathbb{R}^{d\times h}$ is the projection matrix to map the $h$-dimensional hidden state back to $1$-dimensional output $\mathbf{y}_{t,j} \in \mathbb{R} $. $\boldsymbol{\eta}_j \in \mathbb{R}^{h}$ is the $j$-th row of $\boldsymbol{\eta}$.
The output $\boldsymbol{Y}$ from \eqref{eq:mddema} is denoted as $\boldsymbol{Y} \stackrel{\Delta}{=} \mathrm{EMA}(\boldsymbol{X})$.
Because we do not need to explicitly compute $\mathbf{h}^{(j)}_t$ to get the output $\mathbf{y}_{t,j}$, and the time and space complexity is similar to the standard EMA in \eqref{eq:ema}~(see Appendix~\ref{appendix:ema} for the details).
Experimental improvements demonstrate its effectiveness (\S\ref{sec:experiments}).

\begin{figure}[!t]
\centering
\includegraphics[width=0.99\textwidth]{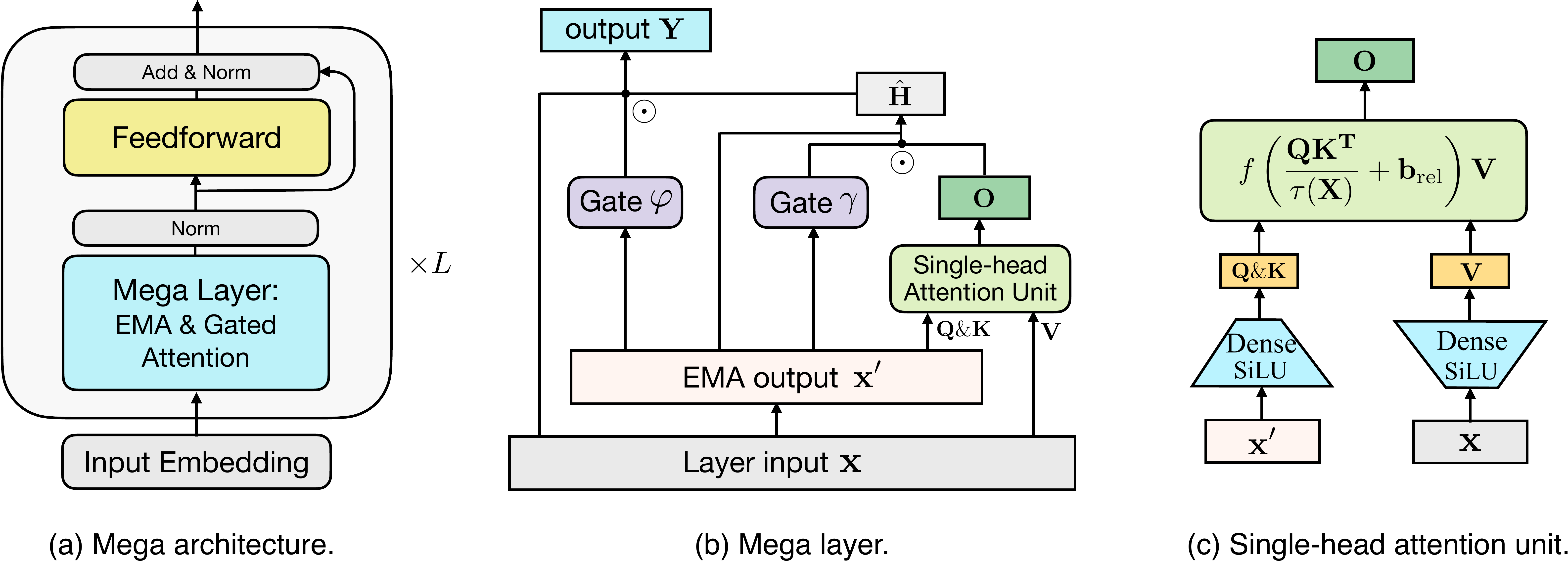}
\caption{\textsc{Mega} -- model architecture. Figure (a) shows the overall architecture of each \textsc{Mega} block. Figure (b) illustrates the gated attention sub-layer equipped with EMA, while Figure (c) displays the details of a single-head attention unit. }
\label{fig:arch}
\end{figure}

\subsection{Moving Average Equipped Gated Attention}
\label{subsec:mega}
The gated attention mechanism in \textsc{Mega} adopts the Gated Recurrent Unit (GRU; \citet{cho2014properties}) and Gated Attention Unit (GAU; \citet{hua2022transformer}) as the backbone architectures, with an EMA-based sub-layer embedded into the calculation of the attention matrix. 
Formally, we first use the output from \eqref{eq:mddema} to compute the shared representation in GAU:
\begin{align}
\boldsymbol{X}' & = \mathrm{EMA}(\boldsymbol{X}) \qquad & \qquad \in \mathbb{R}^{n\times d} \\
\boldsymbol{Z} & = \phi_{\mathrm{silu}}(\boldsymbol{X}' W_z + b_z) \qquad & \qquad \in \mathbb{R}^{n\times z} \label{eq:z}
\end{align}
where $\boldsymbol{X}'$ can be regarded as the updated or contextual input, because it encodes contextual information through EMA. 
$\boldsymbol{Z}$ is the shared representation with $z$ dimensions, with projection matrix $W_z \in \mathbb{R}^{d\times z}$ and bias term $b_z \in \mathbb{R}^{z}$.  
$\phi_{\mathrm{silu}}$ is the self-gated activation function (SiLU)~\citep{ramachandran2017swish,elfwing2018sigmoid}.
Following GAU, the query and key sequences are computed by applying per-dimension scalars and offsets to $\boldsymbol{Z}$, and the value sequence is from the original $\boldsymbol{X}$:
\begin{align}
\boldsymbol{Q} & = \boldsymbol{\kappa}_q \odot \boldsymbol{Z} + \boldsymbol{\mu}_q \qquad & \qquad \in \mathbb{R}^{n\times z} \label{eq:query} \\
\boldsymbol{K} & = \boldsymbol{\kappa}_k \odot \boldsymbol{Z} + \boldsymbol{\mu}_k \qquad & \qquad \in \mathbb{R}^{n\times z} \\
\boldsymbol{V} & = \phi_{\mathrm{silu}}(\boldsymbol{X} W_v + b_v) \qquad & \quad \qquad \in \mathbb{R}^{n\times v} \label{eq:value}
\end{align}
where $\boldsymbol{\kappa}_q$, $\boldsymbol{\mu}_q$, $\boldsymbol{\kappa}_k$, $\boldsymbol{\mu}_k \in \mathbb{R}^{z}$ are the learnable scalars and offsets of queries and keys, respectively.
$v$ is the expanded intermediate dimension for the value sequence. 
The output of attention is computed as follows:
\begin{align}\label{eq:attention2}
\boldsymbol{O} & = f \left(\frac{\boldsymbol{Q}{\boldsymbol{K}}^{T}}{\tau(\boldsymbol{X})} + \boldsymbol{b}_{\mathrm{rel}} \right) \boldsymbol{V} \quad & \quad \qquad \in \mathbb{R}^{n\times v}
\end{align}
The graphical specification is displayed in Figure~\ref{fig:arch} (c).
$\boldsymbol{b}_{\mathrm{rel}} \in \mathbb{R}^{n\times n}$ is the relative positional bias. 
We choose $\boldsymbol{b}_{\mathrm{rel}}$ from existing approaches, including T5~\citep{raffel2020exploring}, RoPE~\citep{su2021roformer}, TUPE~\citep{ke2020rethinking} and ALiBi~\citep{press2021train}. 

Subsequently, \textsc{Mega} introduces the reset gate $\boldsymbol{\gamma}$, the update gate $\boldsymbol{\varphi}$, and computes the candidate activation output $\boldsymbol{\hat{H}}$:
\begin{align}
\boldsymbol{\gamma} & = \phi_{\mathrm{silu}}(\boldsymbol{X}' W_\gamma + b_\gamma) & \in \mathbb{R}^{n\times v} \label{eq:reset} \\
\boldsymbol{\varphi} & = \phi_{\mathrm{sigmoid}}(\boldsymbol{X}' W_\varphi + b_\varphi) & \in \mathbb{R}^{n\times d} \label{eq:update} \\
\boldsymbol{\hat{H}} & = \phi_{\mathrm{silu}}(\boldsymbol{X}' W_h + (\boldsymbol{\gamma} \odot \boldsymbol{O}) U_{h} + b_h) & \in \mathbb{R}^{n\times d} \label{eq:candidate}
\end{align}
The final output $\boldsymbol{Y}$ is computed with the update gate $\boldsymbol{\varphi}$:
% between the input $\boldsymbol{X}$ and the candidate output $\boldsymbol{\hat{H}}$:
\begin{align}
\boldsymbol{Y} & = \boldsymbol{\varphi} \odot \boldsymbol{\hat{H}} + (1 - \boldsymbol{\varphi}) \odot \boldsymbol{X} \qquad & \qquad \in \mathbb{R}^{n\times d} \label{eq:output}
\end{align}
The graphical architecture of a \textsc{Mega} sub-layer is visualized in Figure~\ref{fig:arch} (b).

\paragraph{Laplace Attention Function.}
As mentioned in Section~\ref{subsec:attention}, the softmax function is the most common choice for the attention function $f(\cdot)$.
\citet{so2021searching} recently introduced the squared ReLU function $f_{\mathrm{relu^2}}(\cdot)$ via architecture search techniques, which has shown faster convergence speed and competitive generalization performance on language tasks~\citep{hua2022transformer}.
However, one issue of $f_{\mathrm{relu^2}}(\cdot)$ is that neither its range nor its gradient is bounded, leading to unstable model training (see Appendix~\ref{appendix:laplacevsrelu2} for details). 
To address this issue, we propose a new attention function based on the Laplace function:
\begin{equation}
\label{eq:laplace}
f_{\mathrm{laplace}}(x; \mu,\sigma) = 0.5 \times \left[1 + \mathrm{erf}(\frac{x - \mu}{\sigma\sqrt{2}}) \right]
\end{equation}
where $\mathrm{erf}(\cdot)$ is the error function. 
$\mu$ and $\sigma$ are two coefficients that we adjust to approximate $f_{\mathrm{relu^2}}$, yielding $\mu=\sqrt{1/2}$ and $\sigma=\sqrt{1/4\pi}$.
The derivations and visualization of the Laplace function are provided in Appendix~\ref{appendix:laplace}.

\paragraph{Relation to and Differences from GRU, Flash and S4.}
The computation of the the reset gate $\boldsymbol{\gamma}$, the update gate $\boldsymbol{\varphi}$, and the candidate activation output $\boldsymbol{\hat{H}}$ in (\ref{eq:reset}-\ref{eq:candidate}) is reminiscent of GRUs~\citep{cho2014properties}. 
The main difference is that in a GRU the two gates are applied between the hidden states of the current and previous timesteps, while in \textsc{Mega} they are applied between the outputs from EMA and gated attention sub-layers.
In addition, the output gating mechanism in \eqref{eq:output} is similar to the gated residual connection proposed in \citet{parisotto2020stabilizing,xu2020transformer} to reduce the variance of output $\boldsymbol{Y}$.

The computation of the shared representation $\boldsymbol{Z}$, together with the sequences of queries, keys and values in (\ref{eq:z}-\ref{eq:value}) are inspired from GAU in Flash~\citep{hua2022transformer}.
\textsc{Mega} integrates EMA into GAU by computing $\boldsymbol{Z}$ in \eqref{eq:z} from the EMA output $\boldsymbol{X}'$ rather than the original input $\boldsymbol{X}$, and combining the GAU output with $\boldsymbol{X}'$ for the candidate activation $\boldsymbol{\hat{H}}$ in~\eqref{eq:candidate}. 
Experimental gains over Flash demonstrate the effectiveness of this design chice (\S\ref{subsec:lra}).
% \textsc{Mega} (\S\ref{subsec:lra}).

The multi-dimensional damped EMA can be seen as a simplified variant of a state space model.
From this perspective, \textsc{Mega} is also closely related to S4~\citep{gu2022efficiently}, a state space model with structured state matrices.
S4 leverages the HiPPO framework~\citep{gu2020hippo} to initialize its low-rank structured state matrices, and the computation of the convolutional kernel in S4 requires complex fast Fourier transformers (FFTs).
The EMA sub-layer in \textsc{Mega} applies diagonalization on the state matrix and restricts the diagonal elements in the range of $(0, 1)$.
Thus, the convolution kernel would be a Vandermonde product, which can be computed in an efficient and numerically stable way.
Similar diagonalization has been used in a concurrent work S4D~\citep{gu2022parameterization}.
Moreover, unlike S4 and S4D, the parameter initialization in \textsc{Mega} does not rely on the HiPPO framework.

\subsection{Theoretical Justification of Single-head Gated Attention}
\label{subsec:theory}
Single-head gated attention has been empirically shown as performant as vanilla multi-head attention~\cite{liu2021pay,hua2022transformer}, without any discussions on its theoretical insights.
In this section, we provide theoretical justifications of the expressiveness of single-head gated attention.
To facilitate subsequent analysis, we simplify the notations of the multi-head attention.
Specifically, we denote the sequences of queries, keys and values as the outputs of three transformations of the input sequence:
\begin{equation}
\boldsymbol{Q} = \mathcal{Q}(\boldsymbol{X}), \quad \boldsymbol{K} = \mathcal{K}(\boldsymbol{X}), \quad \boldsymbol{V} = \mathcal{V}(\boldsymbol{X})
\end{equation}
where $\mathcal{Q}$, $\mathcal{K}$, $\mathcal{V}$ are three transformations, such as linear projections.
Let $\boldsymbol{q} \in \boldsymbol{Q} = \{\boldsymbol{q}_1,\ldots, \boldsymbol{q}_n \}$ be a single query vector ($\boldsymbol{q} \in \mathbb{R}^d$), and $\boldsymbol{a} = \mathcal{A}(\boldsymbol{q}, \boldsymbol{K})$ denote the corresponding attention weights of $\boldsymbol{q}$, where $\mathcal{A}$ is the attention transformation, i.e. $f(\cdot)$ in \eqref{eq:attention2}.

For multi-head attention, a common implementation is to split the query into $h$ heads across the model dimension: 
\begin{equation}
\boldsymbol{q} = \left[
\begin{array}{c}
\boldsymbol{q}^{(1)} \\
\vdots \\
\boldsymbol{q}^{(h)}
\end{array}
\right]
\end{equation}
where $\boldsymbol{q}^{(i)} \in \mathbb{R}^{d/h}, \mbox{and} \,\, i \in \{1, \ldots, h \}$ is the query of the $i$-th head.
$\boldsymbol{K}$ and $\boldsymbol{V}$ are split in the same way.
The attention weight of the $i$-th head is $\boldsymbol{a}^{(i)} = \mathcal{A}(\boldsymbol{q}^{(i)}, \boldsymbol{K}^{(i)})$. 
Then, the outputs of single-head and multi-head attention are, respectively:
\begin{equation}
\boldsymbol{O}_{\mathrm{SHA}} = \boldsymbol{a}^T \boldsymbol{V} = \left[ 
\begin{array}{c}
\boldsymbol{a}^T \boldsymbol{V}^{(1)} \\
\vdots \\
\boldsymbol{a}^T \boldsymbol{V}^{(h)} \\
\end{array}
\right],
\quad 
\boldsymbol{O}_{\mathrm{MHA}} = \left[ 
\begin{array}{c}
{\boldsymbol{a}^{(1)}}^T \boldsymbol{V}^{(1)} \\
\vdots \\
{\boldsymbol{a}^{(h)}}^T \boldsymbol{V}^{(h)} \\
\end{array}
\right]
\end{equation}
It is straightforward to see that $\boldsymbol{O}_{\mathrm{MHA}}$ is more expressive than $\boldsymbol{O}_{\mathrm{SHA}}$, because $\boldsymbol{O}_{\mathrm{MHA}}$ leverages $h$ sets of attention weights.

In the single-head gated attention, we introduce a gate vector $\boldsymbol{\gamma} = \mathcal{G}(\boldsymbol{X})$ for each $\boldsymbol{q}$, and the output of single-head gated attention is $\boldsymbol{O}_{\mathrm{SHGA}} = \boldsymbol{O}_{\mathrm{SHA}} \odot \boldsymbol{\gamma}$.
The following theorem reveals the equivalence of $\boldsymbol{O}_{\mathrm{SHGA}}$ and $\boldsymbol{O}_{\mathrm{MHA}}$ w.r.t expressiveness (proof in Appendix~\ref{appendix:thm}):
\begin{theorem}\label{thm}
Suppose the transformation $\mathcal{G}$ is a universal approximator. 
Then, for each $\boldsymbol{X}$ there exists $\boldsymbol{\gamma} = \mathcal{G}(\boldsymbol{X})$ such that 
\begin{equation}
\boldsymbol{O}_{\mathrm{SHGA}} = \boldsymbol{O}_{\mathrm{MHA}}
\end{equation}
\end{theorem}
Theorem~\ref{thm} indicates that by simply introducing the gate vector, $\boldsymbol{O}_{\mathrm{SHGA}}$ is as expressive as $\boldsymbol{O}_{\mathrm{MHA}}$. 
In practice, the transformation $\mathcal{G}$ is commonly modeled by a (shallow) neural network, whose universality of approximation has been extensively studied~\citep{hornik1989multilayer,yarotsky2017error,park2020minimum}.

\subsection{\textsc{Mega} Blocks}
\label{subsec:mega_layer}

The \textsc{Mega} layer (moving average equipped gated attention) is used as a drop-in-replacement for regular attention in Transformer.
It is followed by position-wise feed-forward networks (FFNs) and normalization layers to compose one \textsc{Mega} block.
% We incorporate the position-wise feed-forward networks (FFNs) and normalization layers into \textsc{Mega} blocks. 
As the gated residual connection has already been included in \eqref{eq:output}, we omit the original residual connection and directly apply a normalization layer to $\boldsymbol{Y}.$
Concretely, 
\begin{align}
\boldsymbol{Y} & = \mathrm{Norm}(\mathrm{Mega}(\boldsymbol{X})) \nonumber \\
\boldsymbol{Y}' & = \mathrm{Norm}(\mathrm{FFN}(\boldsymbol{Y}) + \boldsymbol{Y})
\end{align}
where $\boldsymbol{Y}'$ is the output of the \textsc{Mega} block.
The overall architecture of a \textsc{Mega} block is shown in Figure~\ref{fig:arch} (a).
In Transformer, the hidden dimension of FFNs is usually set to $d_{\mathrm{FFN}} = 4d$.
To retain a similar model size with each Transformer block, we reduce the hidden dimension of FFN to $d_{\mathrm{FFN}} = 2d$ and set the expanded dimension $v=2d$ for the value sequence in \eqref{eq:value} throughout this paper, unless specified otherwise.

\begin{figure}[t]
\centering
\includegraphics[width=0.99\textwidth]{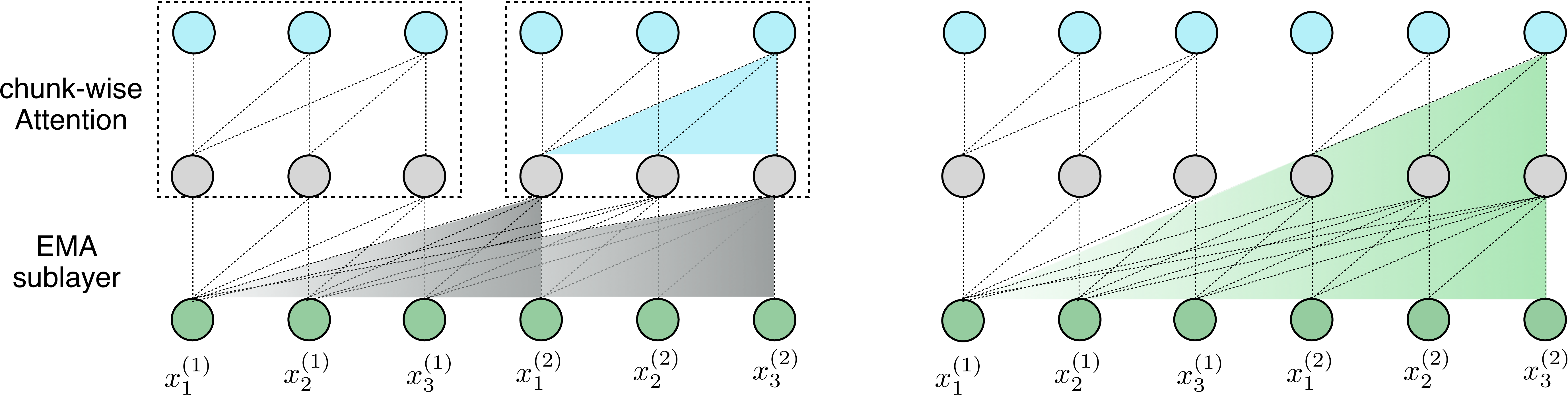}
\caption{Illustration of the \textsc{Mega}-chunk model with two chunks of length 3.}
\label{fig:chunk}
\vspace{-3mm}
\end{figure}

\subsection{\textsc{Mega}-chunk: \textsc{Mega} with Linear Complexity}
\label{subsec:mega-chunk}
So far we have only focused on introducing stronger inductive bias into the attention mechanism, which still has quadratic computational complexity.
% So far our discussion has concentrated on the problem of weak inductive bias in the attention mechanism, without any concerns on the problem of quadratic computational complexity.
In this section, we propose \textsc{Mega}-chunk, a variant of \textsc{Mega} with linear complexity, which simply applies attention to each local chunk of fixed length.

Specifically, we first split the sequences of queries, keys and values in (\ref{eq:query}-\ref{eq:value}) into chunks of length $c$. e.g. $\boldsymbol{Q} = \{\boldsymbol{Q}_1, \ldots,  \boldsymbol{Q}_k\}$, where $k = n/c$ is the number of chunks.\footnote{Keys and values are split in the same way.}
The attention operation in \eqref{eq:attention2} is individually applied to each chunk, yielding linear complexity $O(kc^2)=O(nc)$ w.r.t $n$.
However, this method suffers from the critical limitation of losing contextual information from other chunks.
Fortunately, the EMA sub-layer in \textsc{Mega} mitigates this problem by capturing local contextual information near each token, whose outputs are used as the inputs to the attention sub-layer.
As a result, the effective context being exploited by chunk-wise attention can go beyond the chunk boundary.
Figure~\ref{fig:chunk} illustrates the largest possible dependency length captured by one \textsc{Mega}-chunk block.

\section{Experiments}
\label{sec:experiments}
To evaluate \textsc{Mega}, we conduct experiments on five benchmark sequence modeling tasks across various data types, comparing with current state-of-the-art models on each task.
All the numbers with $\ddag$ indicate results from the baseline models replicated by us.
More detailed descriptions, results and analysis are provided in Appendix~\ref{appendix:experiment}.

\begin{table}
\caption{(\textbf{Long Range Arena}) Accuracy on the full suite of long range arena (LRA) tasks, together with training speed and peak memory consumption comparison on the Text task with input length of 4K. $\ddag$ indicates results replicated by us.}
\label{tab:lra}
\centering
\resizebox{0.99\columnwidth}{!}{
\begin{tabular}{l|cccccc|ccc}
\toprule
\textbf{Models} & \textbf{ListOps} & \textbf{Text} & \textbf{Retrieval} & \textbf{Image} & \textbf{Pathfinder} & \textbf{Path-X} & \textbf{Avg.} & \textbf{Speed} & \textbf{Mem.} \\
\midrule
XFM & 36.37 & 64.27 & 57.46 & 42.44 & 71.40 & \xmark & 54.39 & -- & -- \\
XFM$\ddag$ & 37.11 & 65.21 & 79.14 & 42.94 & 71.83 & \xmark & 59.24 & 1$\times$ & 1$\times$ \\
\midrule
Reformer   & 37.27 & 56.10 & 53.40 & 38.07 & 68.50 & \xmark & 50.67 & 0.8$\times$ & 0.24$\times$ \\
Linformer  & 35.70 & 53.94 & 52.27 & 38.56 & 76.34 & \xmark & 51.36 & 5.5$\times$ & 0.10$\times$ \\
BigBird    & 36.05 & 64.02 & 59.29 & 40.83 & 74.87 & \xmark & 55.01 & 1.1$\times$ & 0.30$\times$ \\
Performer  & 18.01 & 65.40 & 53.82 & 42.77 & 77.05 & \xmark & 51.41 & \textbf{5.7}$\times$ & \textbf{0.11}$\times$ \\
Luna-$256$ & 37.98 & 65.78 & 79.56 & 47.86 & 78.55 & \xmark & 61.95 & 4.9$\times$ & 0.16$\times$ \\
\midrule
S4-v1        & 58.35 & 76.02 & 87.09 & 87.26 & 86.05 & 88.10 & 80.48 & -- & -- \\
S4-v2        & 59.60 & 86.82 & 90.90 & 88.65 & 94.20 & 96.35 & 86.09 & -- & -- \\
S4-v2$\ddag$ & 59.10 & 86.53 & 90.94 & 88.48 & 94.01 & 96.07 & 85.86 & 4.8$\times$ & 0.14$\times$ \\
\midrule
\textsc{Mega} & \textbf{63.14} & \textbf{90.43} & \textbf{91.25} & \textbf{90.44} & \textbf{96.01} & \textbf{97.98} & \textbf{88.21} & 2.9$\times$ & 0.31$\times$ \\
\textsc{Mega}-chunk & 58.76 & 90.19 & 90.97 & 85.80 & 94.41 & 93.81 & 85.66 & 5.5$\times$ & 0.13$\times$ \\
\bottomrule
\end{tabular}
}
\end{table}
\subsection{Long-Context Sequence Modeling}
\label{subsec:lra}
We begin our experiments with an evaluation on the Long Range Arena (LRA) benchmark recently introduced by~\citet{tay2021long}, which is designed for the purpose of evaluating sequence models under the long-context scenario.
They collect six tasks in this benchmark which are ListOps~\citep{nangia2018listops}, byte-level text classification (Text;~\citet{maas2011learning}), byte-level document retrieval (Retrieval; ~\citet{radev2013acl}), image classification on sequences of pixels (Image;~\citet{krizhevsky2009learning}), Pathfinder~\citep{drew2018advances} and its extreme long version (Path-X;~\citet{tay2021long}). 
These tasks consist of input sequences ranging from 1K to 16K tokens and span across a variety of data types and modalities.

Table~\ref{tab:lra} compares \textsc{Mega} against several baselines, including Transformer and its efficient variants, and the state-of-the-art S4 models (both version 1~\citep{gu2022efficiently} and version 2~\citep{gu2022parameterization}).\footnote{The S4-v2 used larger model sizes and better-tuned hyper-parameters than S4-v1. Note that our \textsc{Mega} has similar model size with S4-v1 on each task. We have also experimented with SRU++~\citep{lei2021attention} on Pathfinder but failed to converge on this dataset after tuning hyperparameters.}
To ensure fair comparison, we adjust the number of layers and model dimensions on each task so that \textsc{Mega} has similar number of parameters with S4-v1.
For each experiment, we report the average over 5 runs with different random seeds.
The tuning information and the model details are provided
in the Appendix~\ref{appendix:lra}.

On all the six tasks, \textsc{Mega} substantially outperforms all the baselines. 
We also evaluate \textsc{Mega}-chunk on each task, by setting the chunk size $c=128$ for all the tasks, except Path-X where $c=4096$.
We observe that \textsc{Mega}-chunk consistently performs well, particularly on the three language tasks.
We also examine the speed and memory efficiency of \textsc{Mega} on the byte-level classification task with the input length of 4K.
\textsc{Mega}-chunk is highly efficient, which is about $5.5$ times faster and consumes only $13$\% as much memory as the vanilla Transformer.
It is interesting to see that \textsc{Mega} with full attention field is also much more efficient than Transformer, benefiting from single-head gated attention.

\begin{figure}[!t]
%\begin{minipage}{\textwidth}
\begin{minipage}{.65\textwidth}
    \begin{figure}[H]
        \centering
        \begin{subfigure}[b]{0.49\columnwidth}        	\includegraphics[width=\columnwidth]{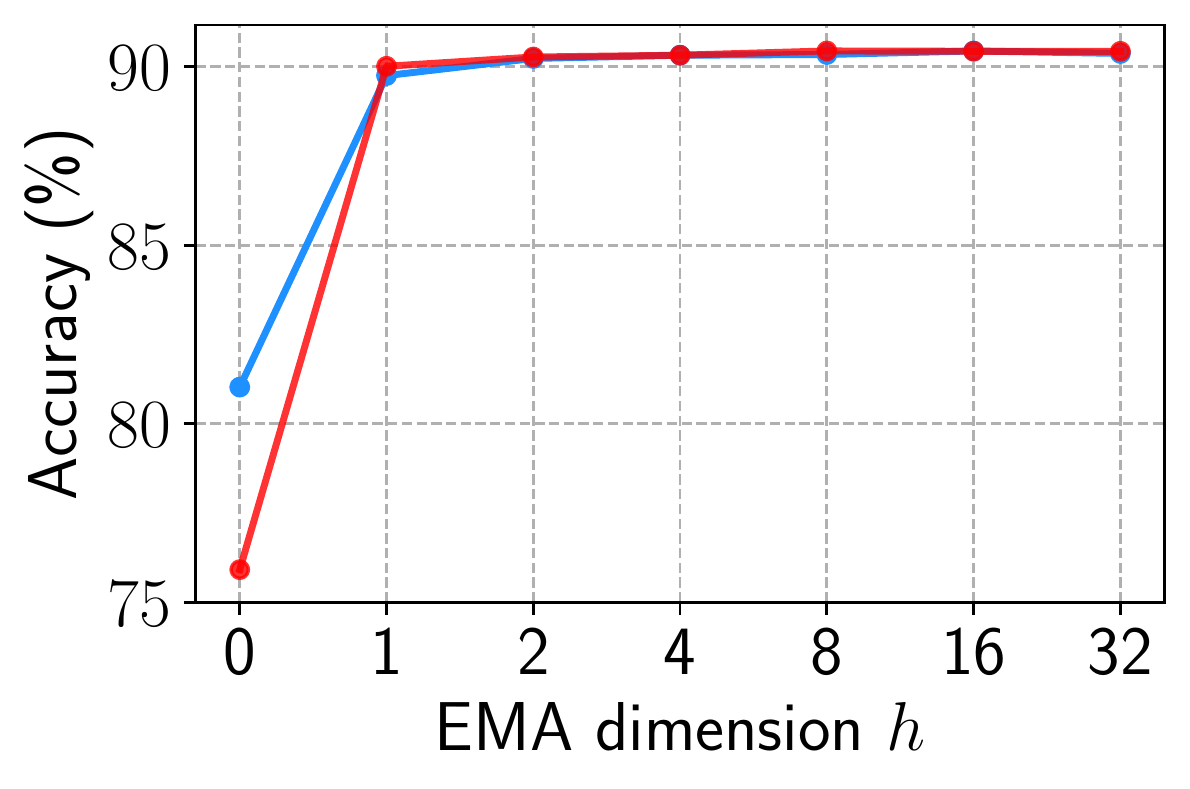}
        	\label{fig:res-xsum-overview}
        \end{subfigure}
        % \hfill
        \begin{subfigure}[b]{0.49\columnwidth}
    	\includegraphics[width=\columnwidth]{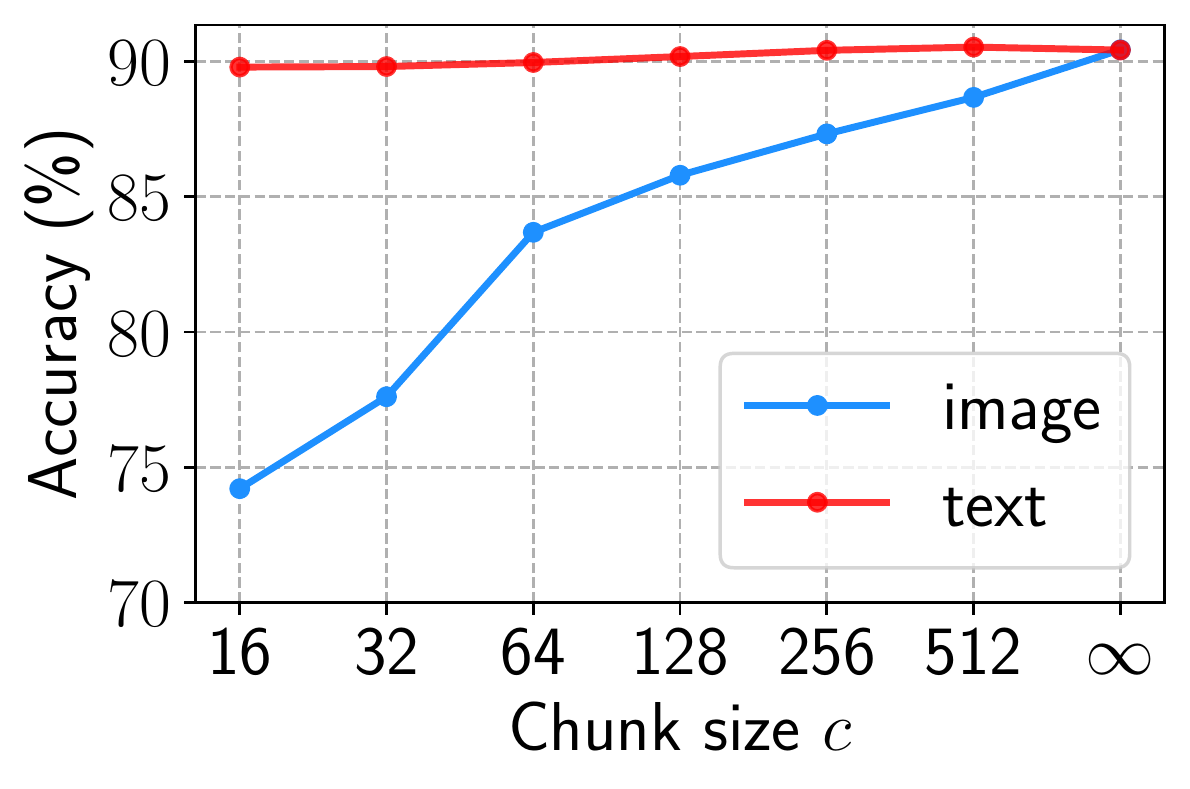}
    	\label{fig:res-mt-overview}
        \end{subfigure}
        \vspace{-12mm}
        \caption{Ablations on EMA dimension and chunk size.\label{fig:hc}}
    \end{figure}
\end{minipage}
\hfill
\begin{minipage}{0.33\textwidth}
\centering
\vspace{6.5mm}
%\resizebox{0.99\columnwidth}{!}{
\begin{tabular}[t]{l|cc}
\toprule
 & Text & Image \\
\midrule
softmax  & \textbf{90.43} & 89.87 \\
relu$^2$ & 90.08 & 90.22 \\
laplace  & 90.22 & \textbf{90.43} \\
\bottomrule
\end{tabular}
%}
\vspace{3mm}
\captionof{table}{Attention functions.}
\label{tab:attn_fn}
\end{minipage}
%\end{minipage}
\vspace{-3mm}
\end{figure}

\paragraph{Analysis of Multi-dimensional Damped EMA.}
To demonstrate the effectiveness of the multi-dimensional damped EMA component in \textsc{Mega}, we performs ablation studies on two LRA tasks --- byte-level text classification (Text) and image classification on sequences of pixels (Image). 
We train \textsc{Mega} models with EMA dimension $h \in \{0, 1, 2, 4, 8, 16, 32\}$, where $h=0$ indicates removing the EMA component.
From the left figure in Figure~\ref{fig:hc}, we see that without the EMA component, model accuracy on both the two tasks declines rapidly.
Meanwhile, with a single dimensional EMA ($h=1$), \textsc{Mega} obtains significant improvements, demonstrating the importance of incorporating inductive bias via EMA.

\paragraph{Analysis of Chunk Size.}
We further analyze the impact of chunk size $c$ on the same two tasks, by varying $c \in \{16, 32, 64, 128, 256, 512, \infty\}$, where $\infty$ indicates the original \textsc{Mega} without chunking.
The right figure in Figure~\ref{fig:hc} shows that image data is more sensitive to chunk size than text data.
On the Text task, \textsc{Mega}-chunk with even a small chunk size $c=16$ is able to achieve around 90\% accuracy.
On the Image task, \textsc{Mega}-chunk with $c=16$ achieves around 75\% accuracy, which is still much better than the vanilla Transformer model.

\paragraph{Analysis of Attention Functions.}
Finally, we evaluate performance with different attention functions.
Table~\ref{tab:attn_fn} shows the accuracy of the three attention functions on the same two tasks.
On text data softmax obtains the best accuracy, while on image data it performs the worst.
The laplace function achieves the best accuracy on image data and also competitive result on text data, being consistently better than relu$^2$.
In the following experiments we use softmax for language tasks and laplace for vision and speech ones.

\subsection{Raw Speech Classification}
To evaluate the capability of \textsc{Mega} on the long-range modeling of speech signals, we apply \textsc{Mega} to classify raw speech (with length 16000), rather than using traditional preprocessing (e.g. convert to MFCC features).
Following~\citet{gu2022efficiently}, we perform speech classification on the SC10 subset of the Speech Commands dataset~\citep{warden2018speech}. 
We experiment with the \textsc{Mega}-chunk variant with $c=1000$, since the computation of \textsc{Mega} and Transformer can not fit in GPU memory. 
As shown in Table~\ref{tab:sc}, our \textsc{Mega}-chunk (base) model with 300K parameters
is able to achieve an accuracy of 96.92 that is slightly worse than 97.50 from the state-of-the-art method S4,\footnote{
Our S4 number is obtained by directly running the official S4 code and is a bit worse than the original reported number (98.32), due to different data splits --- the file reading order is not deterministic across machines with \texttt{os.listdir}.
} while by adding 0.18M parameters our \textsc{Mega}-chunk (big) model performs comparably well with S4.

\subsection{Auto-regressive Language Modeling}
We evaluate \textsc{Mega} on two established language modeling benchmarks --- WikiText-103~\citep{merity2017pointer} and enwik8~\citep{hutter2006human}, which are next-token prediction tasks. WikiText-103 is a word-level language modeling dataset containing 103M training tokens from Wikipedia articles. Following previous work~\citep{baevski2018adaptive,dai2019transformer}, we adopt adaptive softmax and input embeddings and use a vocabulary of 260K tokens.
Enwik8 is a character-level language modeling benchmark that has 100M tokens of unprocessed Wikipedia articles and a vocabulary size of about 200. 
At test time, we split the test data into segments and process each segment sequentially.
In Table~\ref{tab:lm}, we compare with previous top-performing models that are designed to take advantage of longer context, including Transformers~\citep{baevski2018adaptive,al2019character} (XFM-adaptive), Transformer-XL~\citep{dai2019transformer} (XFM-XL) and S4~\citep{gu2022efficiently}. 
% For simplicity, we don't incorporate other recent techniques that are shown useful for language modeling, such as memory augmentation~\citep{grave2017improving,khandelwal2019generalization,dai2019transformer} and efficient attention mechanism for handling long context~\citep{tay2020efficient}. 
On both WikiText-103 and enwik8, we obtain very competitive results, outperforming baselines by a large margin while enjoying much faster (9$\times$) inference speed compared to the Transformer model.
\textsc{Mega} can also naturally achieve \textbf{length extrapolation} at inference time to any sequences that are longer than those seen during training due to the recurrent design of the EMA layer. In addition, we can extrapolate to a longer chunk size for \textsc{Mega} attention due to the use of rotary positional embeddings for training~\citep{su2021roformer}.
We describe them in details and provide complete results of using various test-time chunk sizes and segment lengths in Appendix~\ref{appendix:lm}.

\begin{table}[t]
\begin{minipage}{\textwidth}
\begin{minipage}[b]{0.35\textwidth}
\captionof{table}{(\textbf{SC-Raw}) Accuracy on Speech Commands.}
\label{tab:sc}
\centering
\resizebox{\columnwidth}{!}{
\begin{tabular}[t]{l|cc}
\toprule
 & \multicolumn{2}{c}{\textbf{SC-Raw}} \\
\textbf{Model} & \#\textbf{Param.} & \textbf{Acc.} \\
\midrule
XFM & 786K & \xmark \\
S4$\ddag$ & 300K & \textbf{97.50} \\
\textsc{Mega} (base) & 300K & 96.92 \\
\textsc{Mega} (big) & 476K & 97.30 \\
\bottomrule
\end{tabular}
}
\end{minipage}
\hfill
\begin{minipage}[b]{0.62\textwidth}
\captionof{table}{(\textbf{Language Modeling}) PPL ($\downarrow$) on WikiText-103 and bpc ($\downarrow$) on enwik8.}
\label{tab:lm}
\centering
\resizebox{\columnwidth}{!}{
\begin{tabular}[t]{l|ccccc}
\toprule
& \multicolumn{3}{c}{\textbf{WikiText-103}} & \multicolumn{2}{c}{\textbf{enwik8}} \\
\textbf{Model} & \#\textbf{Param.} & \textbf{PPL} & \textbf{Speed} & \#\textbf{Param.} & \textbf{bpc} \\
\midrule
XFM-adaptive & 247M & 18.66 &  5.6k t/s & - & -\\
XFM-XL & 257M & 18.30 & -  & 41M & 1.06 \\
S4 & 249M & 20.95 & - &- &-\\
\textsc{MEGA} & 252M & \textbf{18.07} & 48k t/s & 39M & \textbf{1.02 }\\
\bottomrule
\end{tabular}
}
\end{minipage}
\end{minipage}
\end{table}

\subsection{Neural Machine Translation}
To evaluate \textsc{Mega} on sequence-to-sequence modeling, we conduct experiments on a standard machine translation benchmark, WMT 2016 English-German news translation (WMT'16), consisting of 4.5M sentence pairs of training data.
Following \citet{ott2018scaling}, we validate on \emph{newstest13} and test on \emph{newstest14}.
The \textsc{Mega} models closely follow the architecture of Transformer-base: 6 encoder and decoder layers with model dimension $d=512$.

Table~\ref{tab:mt} presents the BLEU scores on the test sets of WMT'16 from two directions: EN$\rightarrow$DE and DE$\rightarrow$EN.
For each experiment, we report the average of both tokenized and SacreBLEU\footnote{signature: \texttt{nrefs:1|case:mixed|eff:no|tok:13a|smooth:exp|version:1.5.1}}~\citep{post-2018-call} scores with 5 different random seeds.
\textsc{Mega}-base significantly outperforms Transformer-base by over $1.1$ BLEU.
We also report results of \textsc{Mega} with the Laplace attention function, which slightly but consistently underperforms Softmax. 

\subsection{Image Classification}
To evaluate \textsc{Mega} on a large-scale image classification task, we conduct experiments on the Imagenet-$1k$~\citep{deng2009imagenet} dataset, which consists of 1.28M training images and 50K validation images from 1000 classes. 
Top-1 accuracy on the validation set is reported in Table~\ref{tab:imagenet} to assess various models.
\textsc{Mega} obtains about $0.5$\% accuracy improvement over DeiT-B~\citep{touvron2021training}.
We mostly follow DeiT's approach of applying several data augmentation and regularization methods that facilitate the training process, including Cutmix~\citep{yun2019cutmix}, Mixup~\citep{zhang2017mixup}, stochastic depth~\citep{huang2016deep}, repeated augmentation~\citep{hoffer2020augment}, Rand-Augment~\citep{cubuk2020randaugment}, and random erasing~\citep{zhong2020random}.
These methods were highly tuned towards optimizing the performance of DeiT, which might be sub-optimal for \textsc{Mega}. 
Exploring the optimal data augmentation and regularization methods for \textsc{Mega} is an interesting direction for future work. 
More training details are presented in the Appendix~\ref{appendix:imagenet}. 

\begin{table}[t]
\begin{minipage}{\textwidth}
\begin{minipage}[b]{0.49\textwidth}
\captionof{table}{(\textbf{WMT'16)} Test BLEU scores.}
\label{tab:mt}
\centering
\resizebox{\columnwidth}{!}{
\begin{tabular}[t]{l|cccc}
\toprule
 & \multicolumn{2}{c}{\textbf{EN-DE}} & \multicolumn{2}{c}{\textbf{DE-EN}} \\
\textbf{Model} & \textbf{Token.} & \textbf{Sacre.} & \textbf{Token.} & \textbf{Sacre.} \\
\midrule
XFM-base & 27.30 & -- & -- & -- \\
XFM-base$\ddag$ & 27.97 & 27.33 & 31.92 & 31.33 \\
\textsc{Mega}-softmax & \textbf{29.18} & \textbf{28.47} & \textbf{32.90} & \textbf{32.35} \\
\textsc{Mega}-laplace & 28.95 & 28.27 & 32.81 & 32.22 \\
\bottomrule
\end{tabular}
}
\end{minipage}
\hfill
\begin{minipage}[b]{0.48\textwidth}
\captionof{table}{(\textbf{ImageNet}) Top-1 accuracy.}
\label{tab:imagenet}
\centering
\resizebox{\columnwidth}{!}{
\begin{tabular}[t]{l|ccc}
\toprule
\textbf{Model} & \textbf{Img. size} & \#\textbf{Param.} & \textbf{Acc.} \\
\midrule
ResNet-152 & $224^2$ & 60M & 78.3 \\
VIT-B  & $384^2$ & 86M & 77.9 \\
DeiT-B & $224^2$ & 86M & 81.8 \\
\textsc{Mega} & $224^2$ & 90M & \textbf{82.3} \\
\bottomrule
\end{tabular}
}
\end{minipage}
\end{minipage}
\end{table}

\section{Related Work}

A number of techniques have been recently introduced to address the two issues of Transformer models; we only mention a few here due to space limits.
\paragraph{Inductive Bias.}
To incorporate stronger inductive bias into the attention mechanism, one research direction focuses on injecting position information via advanced positional encoding methods, including absolute and relative positional embeddings~\citep{vaswani2017attention,huang2020improve,ke2020rethinking}, and relative positional biases~\citep{su2021roformer,press2021train}.
Another line of research combines the attention mechanism with other neural architectures with intrinsic strong inductive bias, such as convolutional~\citep{gehring2017convolutional,dai2021coatnet} and recurrence~\citep{dai2019transformer,rae2020compressive,lei2021attention}.

\paragraph{Computational Efficiency.}
Many advanced variants of Transformer models (\emph{`xformers'})~\citep{tay2020efficient,tay2021long} have recently emerged to improve the time and memory efficiency. 
Popular techniques include sparse attention patterns~\citep{parmar2018image,beltagy2020longformer,kitaev2020reformer}, low-rank approximations of the attention matrix~\citep{wang2020linformer,ma2021luna}, and approximations through kernelization~\citep{choromanski2020rethinking,peng2021random}.
Although these models demonstrate better \emph{asymptotic} complexity for long sequences, their efficiency gains are less prominent for moderate length sequences and their performance remains behind that of Transformers with regular attention.

\paragraph{Convolutional Neural Networks with Continuous Kernels.}
As EMA and more general state space models such as S4 can be regarded as a convolution transform with kernel size equal to the sequence length, \textsc{Mega} is also relevant with CNNs with continuous kernels, including CKConv~\citep{romero2021ckconv}, FlexConv~\citep{romero2021flexconv} and CCNN~\citep{romero2022towards}.

\section{Conclusion}
We have introduced \textsc{Mega}, a simple, efficient and effective neural architecture used as a drop-in replacement for regular multi-head attention. 
By leveraging the classic exponential moving average (EMA) approach, \textsc{Mega} is capable of incorporating stronger inductive biases into the attention mechanism. 
Moreover, the EMA approach enables the design of \textsc{Mega}-chunk, an efficient variant of \textsc{Mega} with linear complexity. 
On five sequence modeling tasks across various data types, \textsc{Mega} achieves impressive improvements over a variety of strong baselines, including previous state-of-the-art systems.
These improvements lead to a potential direction of future work to apply \textsc{Mega} for multi-modality modeling.

\bibliography{mega}

\newpage
\section*{Appendix: Mega: Moving Average Equipped Gated Attention}
\appendix

\section{Efficient Computation of Multi-dimensional Damped EMA}
\label{appendix:ema}
Note that the computation of the multi-dimensional damped EMAs of different dimensions are entirely independent of each other.
Without loss of generality, we set $d=1$ and omit the dimension index $j$ in the following formulations. 
We denote the initial hidden state as $\boldsymbol{h}_0$.
The multi-dimensional damped EMA defined in \eqref{eq:mddema} can be vectorized into the following formulation:
\begin{align}
\mathbf{h}_t & = \boldsymbol{\alpha} \odot \mathbf{u}_t + (1 - \boldsymbol{\alpha} \odot \boldsymbol{\delta}) \odot \mathbf{h}_{t-1} \\
\mathbf{y}_{t} & = \boldsymbol{\eta}^T \mathbf{h}_t
\end{align}
where $\boldsymbol{\alpha}$, $\boldsymbol{\delta}$, and $\boldsymbol{\eta} \in \mathbb{R}^{h}$.
$\mathbf{u}_t = \boldsymbol{\beta} \mathbf{x}_{t} \in \mathbb{R}^{h}$ and $\mathbf{h}_t \in \mathbb{R}^{h}$ is the EMA hidden state at timestep $t$.

Let's denote $\boldsymbol{\phi} = 1 - \boldsymbol{\alpha} \odot \boldsymbol{\delta}$. Then, unrolling the above two equations explicitly yields:
\begin{align*}
\mathbf{h}_1 & = \boldsymbol{\phi} \odot \mathbf{h}_{0} + \boldsymbol{\alpha} \odot \boldsymbol{\beta} \mathbf{x}_{1} & 
\mathbf{h}_2 & = \boldsymbol{\phi}^2 \odot \mathbf{h}_{0} + \boldsymbol{\phi} \odot \boldsymbol{\alpha} \odot \boldsymbol{\beta} \mathbf{x}_{1} + \boldsymbol{\alpha} \odot \boldsymbol{\beta} \mathbf{x}_{2} & \ldots \\
\mathbf{y}_1 & = \boldsymbol{\eta}^T\boldsymbol{\phi} \odot \mathbf{h}_{0} + \boldsymbol{\eta}^T \boldsymbol{\alpha} \odot \boldsymbol{\beta} \mathbf{x}_{1} & 
\mathbf{y}_2 & = \boldsymbol{\eta}^T\boldsymbol{\phi}^2 \odot \mathbf{h}_{0} + \boldsymbol{\eta}^T\boldsymbol{\phi} \odot \boldsymbol{\alpha} \odot \boldsymbol{\beta} \mathbf{x}_{1} +  \boldsymbol{\eta}^T\boldsymbol{\alpha} \odot \boldsymbol{\beta} \mathbf{x}_{2}  & \ldots
\end{align*}
This can be written into a vectorized formula:
\begin{align}
\mathbf{y}_t & = \boldsymbol{\eta}^T\boldsymbol{\phi}^t \odot \mathbf{h}_{0} + \boldsymbol{\eta}^T\boldsymbol{\phi}^{t-1} \odot \boldsymbol{\alpha} \odot \boldsymbol{\beta} \mathbf{x}_{1} + \ldots +  \boldsymbol{\eta}^T\boldsymbol{\alpha} \odot \boldsymbol{\beta} \mathbf{x}_{t} \\
\mathbf{y} & = \mathcal{K} * \mathbf{x} + \boldsymbol{\eta}^T\boldsymbol{\phi}^t \odot \mathbf{h}_{0} \label{eq:mega:conv}
\end{align}
where $*$ is the convolution transform with kernel $\mathcal{K} \in \mathbb{R}^{n}$:
\begin{equation}
\mathcal{K} = \left(\boldsymbol{\eta}^T(\boldsymbol{\alpha} \odot \boldsymbol{\beta}), \,\, \boldsymbol{\eta}^T(\boldsymbol{\phi} \odot \boldsymbol{\alpha} \odot \boldsymbol{\beta}), \,\, \ldots, \,\, \boldsymbol{\eta}^T(\boldsymbol{\phi}^t \odot \boldsymbol{\alpha} \odot \boldsymbol{\beta}
\right)
\end{equation}
In the proposed multi-dimensional damped EMA, $\mathcal{K}$ can be efficiently computed by the Vandermonde product.
With $K$ provided, the output $\mathbf{y}$ in \eqref{eq:mega:conv} can be computed efficiently with FFTs.

\section{Proof of Theorem 1}
\label{appendix:thm}
\begin{proof}
We split $\boldsymbol{\gamma}$ into $h$ heads in the same way as $\boldsymbol{Q}$, $\boldsymbol{K}$, and $\boldsymbol{V}$:
\begin{displaymath}
\boldsymbol{\gamma} = \left[
\begin{array}{c}
\boldsymbol{\gamma}^{(1)} \\
\vdots \\
\boldsymbol{\gamma}^{(h)}
\end{array}
\right]
\end{displaymath}
Then we have 
\begin{displaymath}
\boldsymbol{O}_{\mathrm{SHGA}} = \boldsymbol{a}^T \boldsymbol{V} \odot \boldsymbol{\gamma} = \left[ 
\begin{array}{c}
\boldsymbol{a}^T \boldsymbol{V}^{(1)} \odot \boldsymbol{\gamma}^{(1)} \\
\vdots \\
\boldsymbol{a}^T \boldsymbol{V}^{(h)} \odot \boldsymbol{\gamma}^{(h)} \\
\end{array}
\right]
\end{displaymath}
To prove Theorem~\ref{thm}, we need to find $\boldsymbol{\gamma}$ such that
\begin{displaymath}
\boldsymbol{a}^T \boldsymbol{V}^{(i)} \odot \boldsymbol{\gamma}^{(i)} = {\boldsymbol{a}^{(i)}}^T \boldsymbol{V}^{(i)} \,\, \Longleftrightarrow \,\, \boldsymbol{\gamma}^{(i)} = {\boldsymbol{a}^{(i)}}^T \boldsymbol{V}^{(i)} \oslash \boldsymbol{a}^T \boldsymbol{V}^{(i)}, \,\, \forall i \in \{1, \ldots, h\},
\end{displaymath}
where $\oslash$ is the element-wise divide operation.
Since $\mathcal{G}(\boldsymbol{X})$ is a universal approximator and $\boldsymbol{Q}$, $\boldsymbol{K}$, $\boldsymbol{V}$ and $\boldsymbol{a}$ are all transformed from $\boldsymbol{X}$, $\boldsymbol{\gamma}$ can theoretically recover ${\boldsymbol{a}^{(i)}}^T \boldsymbol{V}^{(i)} \oslash \boldsymbol{a}^T \boldsymbol{V}^{(i)}, \,\, \forall \boldsymbol{X}$.
\end{proof}

\section{Laplace Attention Function}
\label{appendix:laplace}

\begin{figure}[!h]
\begin{minipage}{\textwidth}
\begin{figure}[H]
\centering
\begin{subfigure}[b]{0.46\columnwidth}        	
\includegraphics[width=\columnwidth]{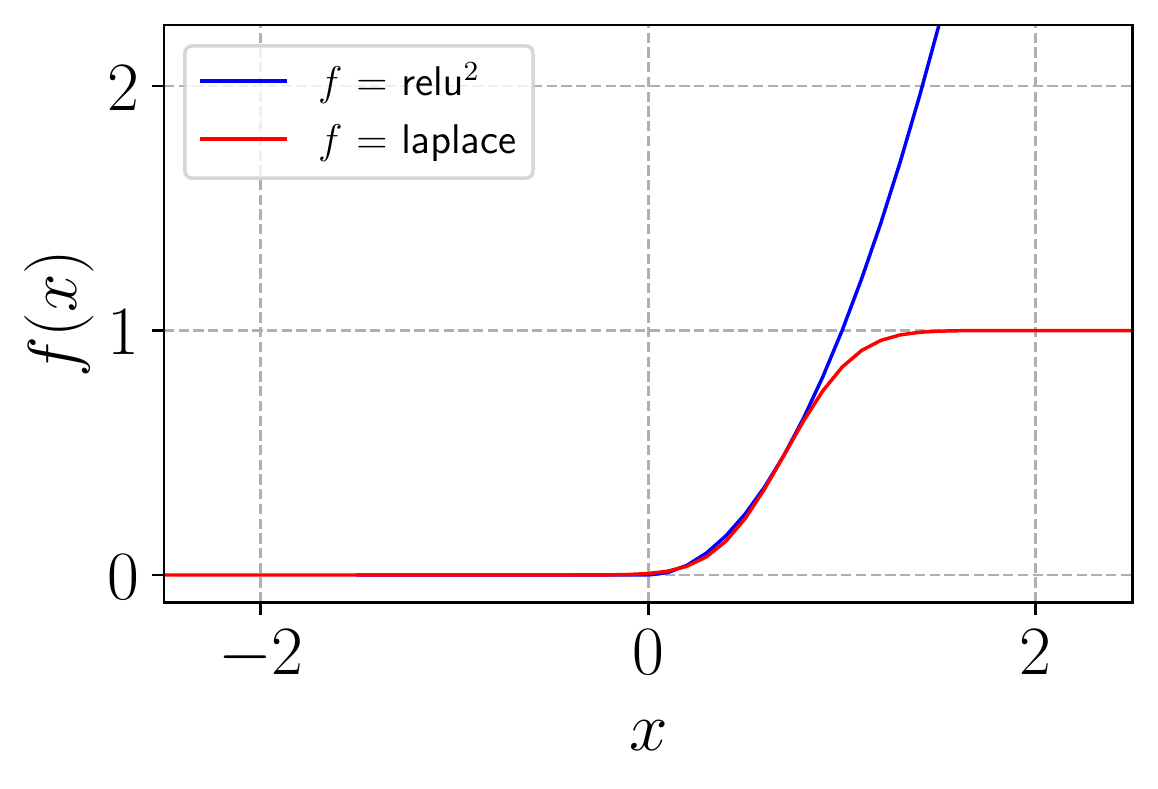}
\label{fig:attn_fn:visualization}
\end{subfigure}
\hfill
\begin{subfigure}[b]{0.5\columnwidth}
\includegraphics[width=\columnwidth]{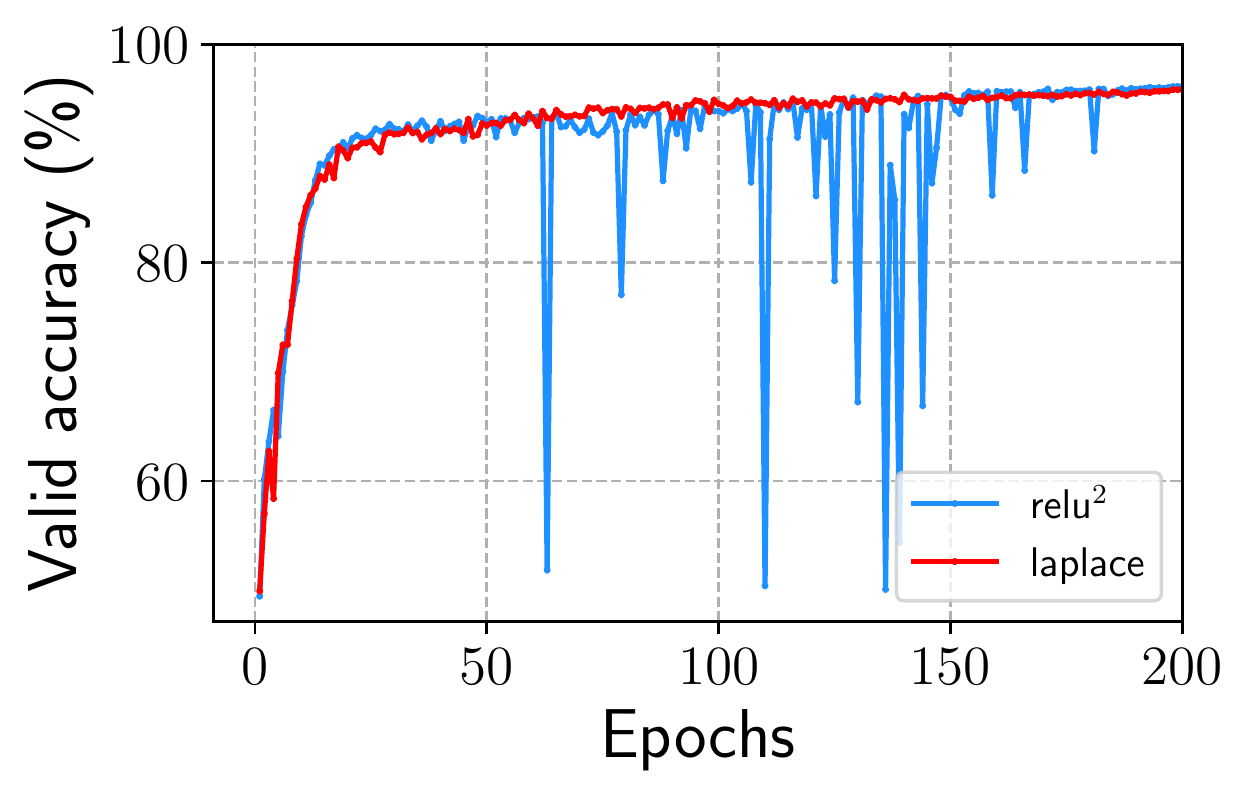}
\label{fig:attn_fn:stability}
\end{subfigure}
\vspace{-6mm}
\caption{Laplace vs. ReLU$^2$. \label{fig:attn_fn}}
\end{figure}
\end{minipage}
\end{figure}

To approximate the squared ReLU function with the Laplace function in \eqref{eq:laplace}, we need to select proper coefficients $\mu$ and $\sigma$.
We derive the values of $\mu$ and $\sigma$ by solving the following two equations at $x=\sqrt{2}$:
\begin{align}
f_{\mathrm{relu2}}(\sqrt{2}) & = f_{\mathrm{laplace}}(\sqrt{2}) \label{eq:laplace1} \\
f'_{\mathrm{relu2}}(\sqrt{2}) & = f'_{\mathrm{laplace}}(\sqrt{2}) \label{eq:laplace2}
\end{align}
The Eq.~\eqref{eq:laplace1} delivers $\mu=\sqrt{1/2}$ and Eq.~\ref{eq:laplace2} subsequently provides $\sigma=\sqrt{1/4\pi}$.
Figure~\ref{fig:attn_fn} visualizes the two functions.

\subsection{Stability: Laplace vs. Squared ReLU}
\label{appendix:laplacevsrelu2}
Besides performance improvements, we also investigate the stability of the two attention functions.
We conduct experiments on the LRA Pathfinder task with \textsc{Mega} models with the two functions.
Figure~\ref{fig:attn_fn} presents the accuracy on the validation set across training epochs.
We observe that Laplace is much more stable than ReLU$^2$.

\newpage

\section{Experimental Details}
\label{appendix:experiment}

\subsection{Long Range Arena (LRA)}
\label{appendix:lra}
For all tasks, we closely follow \citet{tay2020efficient} for details such as data preprocessing, data split, etc.
The hyper-parameters of \textsc{Mega} models on these tasks are listed in Table~\ref{tab:lra:hyps}.
\begin{table}[!h]
\caption{Hyper-parameters of \textsc{Mega} models on LRA and raw speech classification tasks. BSZ is batch size, LR is learning rate and WD is weight decay.
BN, LN and SN refer to Batch Normalization, Layer Normalization and Scale Normalization.}
\label{tab:lra:hyps}
\centering
\resizebox{\columnwidth}{!}{
\begin{tabular}{l|cccccccccccccc}
\toprule
\textbf{Task} & \textbf{Depth} & $d_\mathrm{model}$ & $d_\mathrm{FFN}$ & $z$ & $v$ & $h$ & \textbf{Attn-FN} & \textbf{Norm} & \textbf{Pre-norm} & \textbf{BSZ} & \textbf{LR} & \textbf{Dropout} & \textbf{WD} & \textbf{Epochs} \\
\midrule
\textbf{ListOps}    & 6 & 80  & 160 & 64 & 160 & 16 & softmax & LN & False & 64 & 0.001 & 0.1 & 0.01 & 60 \\
\textbf{Text}       & 4 & 128 & 256 & 64 & 256 & 16 & softmax & SN & False & 50 & 0.004 & 0.1 & 0.01 & 50 \\
\textbf{Retrieval}  & 6 & 128 & 256 & 64 & 256 & 16 & softmax & SN & False & 64 & 0.003 & 0.1 & 0.04 & 40 \\
\textbf{Image}      & 8 & 160 & 320 & 96 & 320 & 16 & laplace & BN & True  & 50 & 0.01  & 0.0 & 0.02 & 200 \\
\textbf{Pathfinder} & 6 & 128 & 256 & 64 & 256 & 16 & laplace & BN & True  & 128 & 0.01  & 0.0 & 0.01 & 200 \\
\textbf{Path-X}     & 4 & 64  & 128 & 32 & 128 & 16 & laplace & BN & True  & 128 & 0.01  & 0.0 & 0.01 & 100 \\
\midrule
\textbf{SC-Raw (base)} & 6 & 60 & 120 & 30 & 120 & 16 & laplace & BN & True & 20 & 0.01 & 0.0 & 0.01 & 200 \\
\textbf{SC-Raw (big)}  & 6 & 72 & 144 & 36 & 144 & 16 & laplace & BN & True & 20 & 0.008 & 0.0 & 0.01 & 200 \\
\bottomrule
\end{tabular}
}
\end{table}

\subsection{Raw Speech Classification}
\label{appendix:sc}
% Speech signals are often sampled at high frequency and long-context sequence modeling is required to work with such signals. Traditional sequence models use extensive preprocessing (e.g. convert to MFCC features) to reduce the length of the input and overcome the long-range modeling challenge. 
% In contrast, we apply \textsc{Mega} to classify raw speech (with length 16000). 
Following~\citet{gu2022efficiently}, we perform speech classification on the SC10 subset of the Speech Commands dataset~\citep{warden2018speech}, which is a 10-class classification task. The chunk size of \textsc{Mega}-chunk is 1000.
Other hyper-parameters are listed in Table~\ref{tab:lra:hyps}.
% \jh{to fill hyperparams}

\subsection{Language Modeling}
\label{appendix:lm}
\paragraph{Training details} We use the data of WikiText-103 and enwik8 and their splits provided by~\citet{dai2019transformer}. At training time, we split the training data into segments; each segment contains $m$ consecutive chunks, where the chunk size is the effective attention length. $m$ is a random integer variable uniformly sampled from $[cl, ch]$. We use $[cl, ch]=[2, 6]$ for WikiText-103 and $[cl, ch]=[2, 4]$ for enwik8. Other training hyperparameters including optimizer, learning rate scheduler and architecture are presented in Table~\ref{tab:lm:hyps}.

\paragraph{Length extrapolation at inference time} We employ \textsc{Mega}-chunk~(\S\ref{subsec:mega-chunk}) for training and set the attention chunk size to be 1024 and 2048 for WikiText-103 and enwik8 respectively. 
To use a longer Mega attention length at inference time than the one used at training time (i.e. 1024 or 2048), we apply rotary positional embedding~\citep{su2021roformer} to the attention sublayer.
At test time, we split the test data into $K$ segments and sequentially process each segment by $m$ chunks, i.e. the maximum context length of each segment is $\frac{\# \text{test tokens}}{K}$.
In Table~\ref{tab:lm}, we report test results that use longer chunk sizes (attention lengths) of 2048 and 4096 for WikiText-103 and enwik8 respectively. 
\textsc{Mega} can naturally extrapolate at inference time to sequences longer than those seen during training due to the recurrent design of the EMA layer. That design enables the inputs of each chunk to access the historic context through EMA as illustrated in Figure~\ref{fig:chunk}.
On the other hand, due to the use of rotary positional embeddings, attention can be performed on longer chunk sizes at test time than those seen during training. We hope these two types of length extrapolation are clear to readers. 
We provide the ablation studies on these two types of length extrapolation below, i.e. extrapolation to longer context by increasing input sequence lengths and extrapolation to longer attention lengths by increasing the chunk size.

\begin{figure}[!t]
\begin{minipage}{.99\textwidth}
\begin{figure}[H]
\centering
\begin{subfigure}[b]{0.49\columnwidth}        	
\includegraphics[width=\columnwidth]{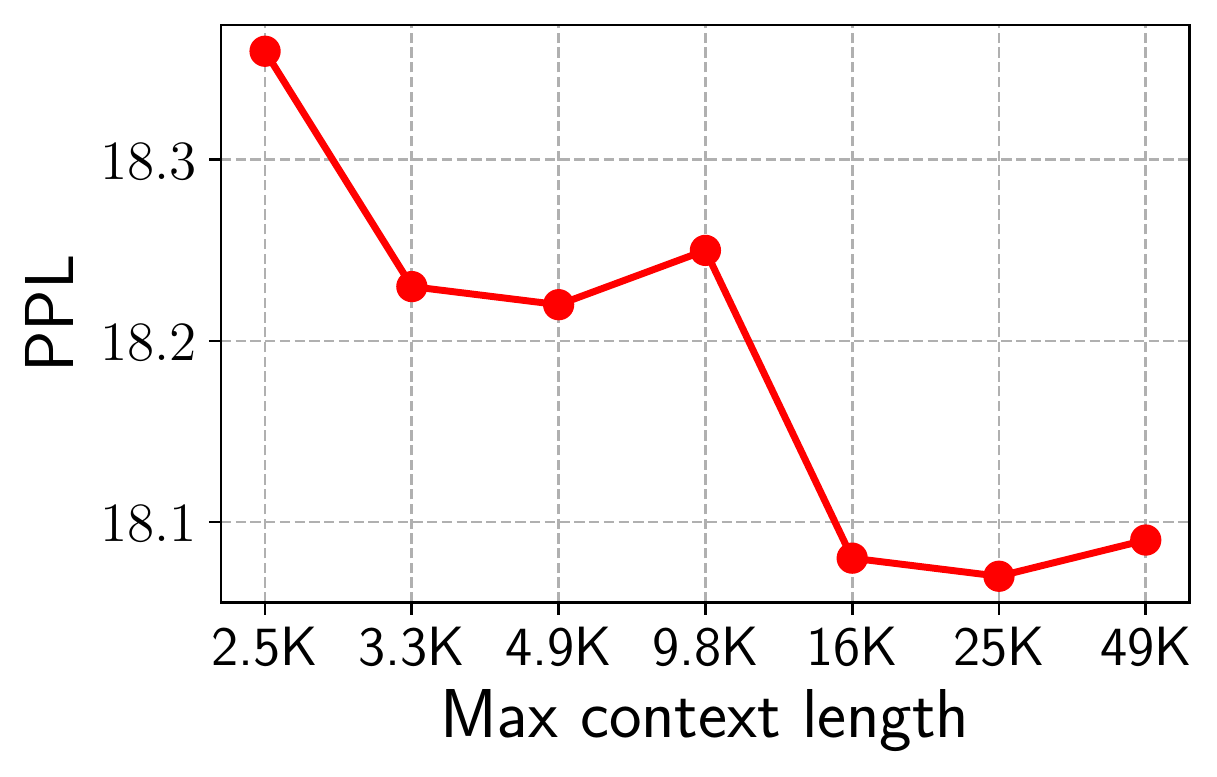}
\end{subfigure}
% \hfill
\begin{subfigure}[b]{0.49\columnwidth}
\includegraphics[width=\columnwidth]{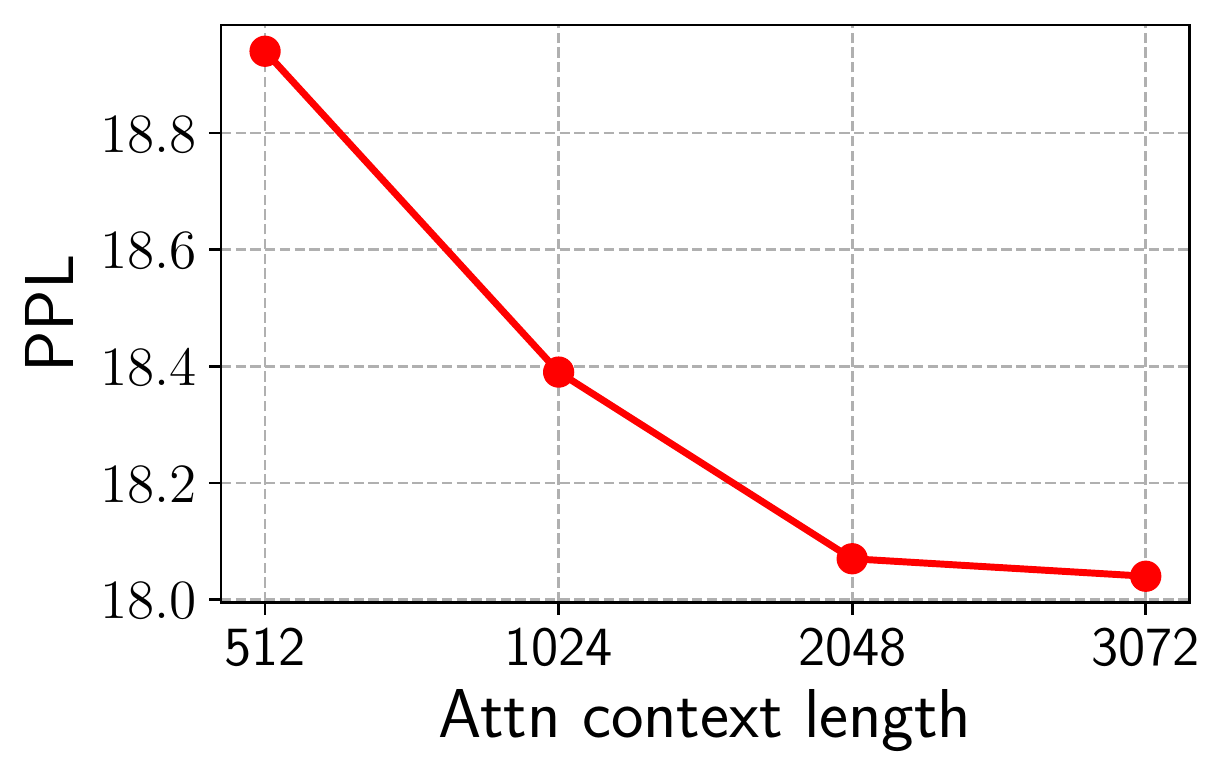}
\end{subfigure}
\caption{Ablation studies of using different context lengths and attention lengths on WikiText-103. \label{fig:length_ppl}}
\end{figure}
\end{minipage}
\end{figure}

\paragraph{Ablations on context lengths}
First, we fix the chunk size to be 2048 and vary $K$ within $[100, 75, 50, 25, 15, 10, 5]$ corresponding to maximum context tokens of $[2.5\text{K}, 3.3\text{K}, 4.9\text{K}, 9.8\text{K}, \\16\text{K}, 25\text{K}, 49\text{K}]$. We plot the test PPL as we increase the context length in the left of Figure~\ref{fig:length_ppl}. Although at training time, the maximum context length the model has seen is 6144, \textsc{Mega} can extrapolate to longer context lengths. The plot shows that 
PPL decreases as the context length is increased and the improvements saturate when the context length is longer than 25K. This is consistent with the observations in~\citet{press2021train}.

\paragraph{Ablations on attention chunk sizes}
Next, we fix the context length to be 25K and increase the chunk size from 512 to 3072. As shown in the right side of Figure~\ref{fig:length_ppl}, \textsc{Mega} consistently improves as we increase the attention length although it only uses an attention length of 1024 during training.
This contradicts with the findings in Alibi~\citep{press2021train}, which finds that rotary embeddings don't generalize to longer lengths and result in higher PPL.
\newpage
\begin{table}[!h]
\caption{Hyper-parameters of models for language modeling.}
\label{tab:lm:hyps}
\centering
\begin{tabular}{l|ccc}
\toprule
 & \textbf{WikiText-103}  & \textbf{enwik8} \\
\midrule
Batch Size $\times$ GPUs  & 6144 $\times$ 24 & 8192 $\times$ 8 \\
Optimizer & AdamW & AdamW \\
Learning Rate &  0.005 & 0.005 \\
Adam-$\beta$ & $(0.9, 0.98)$ & $(0.9, 0.98)$ \\
Learning Rate Decay & linear & linear \\
Weight Decay & 0.1 & 0.1 \\
Dropout & 0.3 & 0.1 \\
Attention Dropout & 0.1 & 0.0 \\
FFN Hidden Dropout & 0.1 & 0.0 \\
Gradient Clipping & 1.0 & 1.0 \\
Warmup steps & 24K & 24K \\
Total updates & 400K & 400K \\
\midrule
Decoder Layers & 16 & 12 \\
Model size & 1024 & 512 \\
FFN Hidden size & 1536 & 1024 \\
Shared Repr. size ($z$) & 256 & 128 \\
Value Seq. size ($v$) & 1536 & 1024 \\
EMA dimension ($h$) & 16 & 16 \\
Chunk size & 1024 & 2048 \\
Total Parameters & 252M & 39M \\
\bottomrule
\end{tabular}
% \vspace{-8pt}
\end{table}

\newpage
\subsection{Machine Translation}
The WMT 2016 English-German dataset contains 4.5M parallel sentence pairs for training.
We following the standard setting~\citep{ott2018scaling}, using Newstest2013 as the validation set and Newstest2014 as the test set.
The dataset is pre-processed following \citep{ma2020apollo}, using the scripts from FairSeq package~\citep{ott2019fairseq}.\footnote{\url{https://github.com/pytorch/fairseq}}
We share the source and target vocabularies within the language pair, with 32K byte pair encoding (BPE) types~\citep{sennrich2016neural}.
The hyper-parameters of Transformer and \textsc{Mega} models are listed in Table~\ref{tab:mt:hyps}.
\begin{table}[!h]
\caption{Hyper-parameters of models for machine translation.}
\label{tab:mt:hyps}
\centering
\begin{tabular}{l|ccc}
\toprule
 & \textbf{XFM}-Base  & \textbf{\textsc{Mega}}-Base \\
\midrule
Batch Size $\times$ GPUs & 8192 $\times$ 8 & 8192 $\times$ 8 \\
Optimizer & AdamW & AdamW \\
Learning Rate &  0.0005 & 0.001 \\
Adam-$\beta$ & $(0.9, 0.98)$ & $(0.9, 0.98)$ \\
Learning Rate Decay & inv. sqrt & linear \\
Weight Decay & $1e-4$ & 0.05 \\
Dropout & 0.1 & 0.15 \\
Attention Dropout & 0.1 & 0.1 \\
FFN Hidden Dropout & 0.1 & 0.1 \\
Gradient Clipping & 1.0 & 1.0 \\
Label Smoothing & 0.1 & 0.1 \\
Warmup steps & 4K & 4K \\
Total updates & 500K & 500K \\ 
\midrule
Encoder Layers & 6 & 6 \\
Decoder Layers & 6 & 6 \\
Model dimension & 512 & 512 \\
FFN Hidden dimension & 2048 & 1024 \\
Shared Repr. dimension ($z$) & -- & 128 \\
Value Seq. dimension ($v$) & -- & 1024 \\
EMA dimension ($h$) & -- & 16 \\
Total Parameters & 65M & 67M \\
\bottomrule
\end{tabular}
% \vspace{-8pt}
\end{table}

\newpage
\subsection{Image Classification}
\label{appendix:imagenet}
Hyper-parameters are listed in Table~\ref{tab:imagenet:hyps}. 
We closely follow \citet{touvron2021training} by reusing most of the their hyper-parameters. 
\begin{table}[!h]
\caption{Ingredients and hyper-parameters of DeiT and \textsc{Mega}.}
\label{tab:imagenet:hyps}
\centering
%\resizebox{0.9\columnwidth}{!}{
\begin{tabular}{lcc}
\toprule
 & DeiT-B & \textsc{Mega} \\
\midrule
Batch size & 1024 & 1024\\
Optimizer & AdamW & AdamW\\
learning rate & 0.001 & 0.002  \\
Learning rate decay & cosine & cosine  \\
Weight decay & 0.05 & 0.05    \\
Epochs & 300 & 300 \\
Warmup epochs & 5 & 20 \\
\midrule
Label smoothing & 0.1 & 0.1 \\
Dropout & \xmark & \xmark \\
Stoch. Depth & 0.1 & 0.2 \\
Repeated Aug & 3 & 3 \\
Gradient Clip. & \xmark & 1.0 \\
\midrule
Rand Augment & 9/0.5 & 9/0.5 \\
Mixup prob. & 0.8 & 0.8 \\
Cutmix prob. & 1.0 & 1.0 \\
Erasing prob. & 0.25 & 0.25 \\
\midrule
Num. Layers & 12 & 12 \\
Model size & 768 & 768 \\
FFN Hidden size & 3072 & 1536 \\
Shared Repr. size ($z$) & -- & 256 \\
Value Seq. size ($v$) & -- & 1536 \\
Total Parameters & 86M & 90M \\
\bottomrule
\end{tabular}
%}
\end{table}

\end{document}